\documentclass[letterpaper, 10 pt, conference]{ieeeconf}  

\usepackage[utf8]{inputenc}
\usepackage[usenames, dvipsnames]{color}
\usepackage{graphicx}
\usepackage{enumerate}
\usepackage{multirow}
\usepackage{graphics}
\usepackage{svg}
\usepackage{graphicx}

\usepackage{amsmath}
\usepackage{nth}
\usepackage{makecell}
\usepackage{colortbl}

\usepackage{tikz}
\usetikzlibrary{shapes,arrows}
\usepackage{subcaption}
\usepackage{amsmath,amsthm,amssymb,mathrsfs} 
\usepackage[final]{pdfpages}
\usepackage[]{placeins,flafter}
\usepackage[none]{hyphenat} 
\usepackage{xcolor}
\usepackage{adjustbox}
\usepackage{balance}
\usepackage{flushend}
\usepackage{comment}
\usepackage{nicefrac}
\usepackage{url}
\usepackage{hyperref}
\usepackage{float}
\usepackage{multicol}
\newcommand{\etal}{\textit{et al.}}

\IEEEoverridecommandlockouts                              

\title{\LARGE \bf
Danish Airs and Grounds: \\ A Dataset for Aerial-to-Street-Level Place Recognition and Localization
}

\author{Andrea Vallone$^{1*}$, Frederik Warburg$^{1*}$, Hans Hansen$^{2}$, Søren Hauberg$^{1}$ and Javier Civera$^{3}$
\thanks{$^{*}$ Equal contribution.}%
\thanks{$^{1}$Technical University of Denmark,}%
\thanks{\ \ {\tt\small {s192327,frwa,sohau}@dtu.dk}}%
\thanks{$^{2}$Dansk Drone Kompagni ApS, {\tt\small hans@dronekompagniet.dk}}%
\thanks{$^{3}$I3A, Universidad de Zaragoza, {\tt\small jcivera@unizar.es}}%
}

\begin{document}

\maketitle

\begin{abstract}
Place recognition and visual localization are particularly challenging in wide baseline configurations. In this paper, we contribute with the \emph{Danish Airs and Grounds} (DAG) dataset, a large collection of street-level and aerial images targeting such cases. Its main challenge lies in the extreme viewing-angle difference between query and reference images with consequent changes in illumination and perspective. The dataset is larger and more diverse than current publicly available data, including more than 50 km of road in urban, suburban and rural areas. All images are associated with accurate 6-DoF metadata that allows the benchmarking of visual localization methods.

We also propose a map-to-image re-localization pipeline, that first estimates a dense 3D reconstruction from the aerial images and then matches query street-level images to street-level renderings of the 3D model. The dataset can be downloaded at: \hyperlink{url{https://frederikwarburg.github.io/DAG/}}{\url{https://frederikwarburg.github.io/DAG/}}.

\end{abstract}

\section{Introduction}

Estimating the 6-Degrees-of-Freedom (6-DoF) camera pose in a known map representation of a scene is a core component in many applications such as autonomous driving, robotics, and augmented reality. Visual localization pipelines are usually divided into two stages. First, a place recognition method obtains a coarse camera pose by finding images from the same place as a given query image among a large database of geo-registered images. Second, a visual localization method estimates an accurate camera pose between the retrieved image and the query image, in most cases relying on feature extraction and matching. 

Handcrafted descriptors (\emph{e.g.}, \cite{cummins2008fab,galvez2012bags}) have shown impressive performance for both place recognition and visual localization, but are limited to small changes in perspective, illumination and scene structure. In the last decade, learning-based feature extractors and descriptors have overcome these limitations, even for drastic appearance changes such as day-to-night or summer-to-winter. The need for training data and fair benchmarking have motivated the release of many large and challenging place recognition~\cite{nordland, torii201524, warburg2020mapillary} and localization~\cite{RobotCarDatasetIJRR, Sattler2018CVPR, eth_ms_visloc_2021} datasets that focus especially on appearance and viewpoint changes. Following this trend, aerial mapping is a particularly interesting application to study viewpoint invariances.

\begin{figure}[t!]
    \centering
    \includegraphics[width=\linewidth]{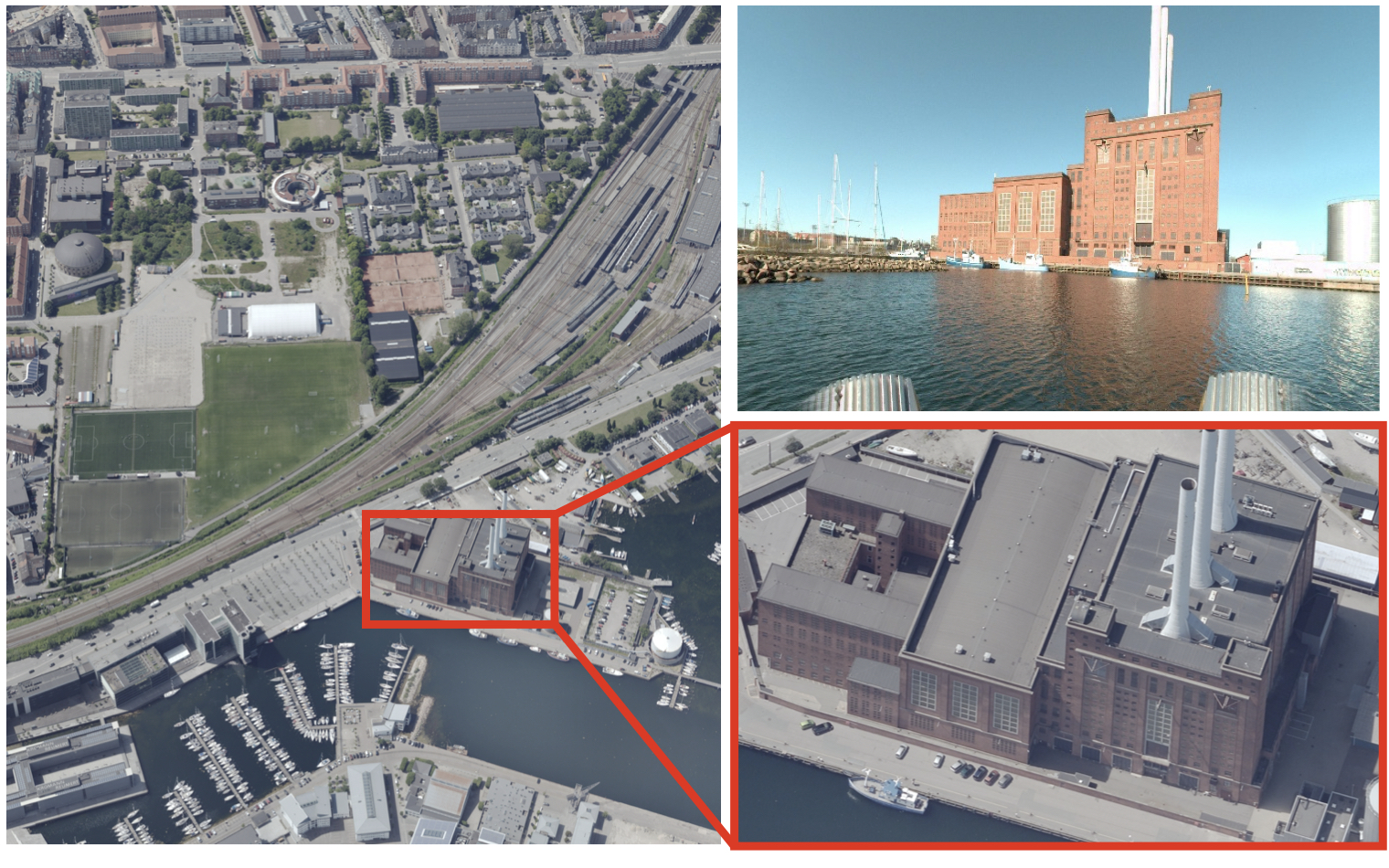}
    \caption{Sample images from our dataset that illustrate the challenge of retrieving the closest database aerial image given a street-level query image, and of registering the query in the aerial reference frame.}
    \vspace{-0.5cm}
    \label{fig:teaser_fig}
\end{figure}

Moreover, aerial mapping has a wide range of applications. Compared to street-view mapping, in which drivers or pedestrians have to traverse every road, aerial images provide a more scalable method for mapping large areas. The alignment of multiple mapping sequences is simpler with airplane photos than street-level sequences, because of the large receptive field and overlap of aerial images. Compared to satellite photos, airplane photos provide oblique views and higher resolution that allows for detailed mapping of building facades (\emph{e.g.} see the detailed texture on the facade of the power plant in Fig. \ref{fig:teaser_fig}). 

This paper contributes to the ongoing research on visual place recognition and localization with a challenging dataset presenting extreme viewpoint changes. Specifically, the \emph{Danish Airs and Grounds} (DAG) dataset targets visual place recognition and localization between aerial and street-level images. DAG contains diverse urban and suburban environments and is currently the largest and most diverse dataset of its kind.

To validate the dataset, we present a pipeline for aerial-to-street-level visual localization. We first create a 3D model from aerial images from which we render street-level images, thus reducing the view-angle difference between query and database images. We show that pre-trained feature descriptors are effective for visual localization between the rendered and the query images. Our pipeline, however, comes at the expense of an expensive 3D reconstruction and rendering process. We hope that the release of the DAG dataset will facilitate research in direct visual localization between aerial and street-level images without the need of rendered views, and that our pipeline will contribute as a valuable baseline method.



\section{Related Work}

The visual image localization pipeline is typically divided into place recognition and $6$-DoF localization. In this section, we will review the main trends of both stages followed by the most common datasets for visual localization.

\subsection{Visual Place Recognition} 

Visual place recognition is often cast as an image retrieval task, where the goal is to find images from the same place as a query image in a large database of geo-registered images. The definition of same-place varies, but usually two places are considered the same if they are within a certain distance radius ($25$ meters is a common choice). Retrieval methods are more scalable than full $6$-DoF motion estimation, but only provide a coarse localization (that of the closest database image). Therefore, place recognition methods are often used as an initial step to constrain the $6$-DoF localization to a few images. \looseness=-1

Classical visual place representations consist of handcrafted local descriptors aggregated with either Bag-of-Words (BoW)~\cite{sivic2003video}, Fischer vectors~\cite{perronnin2010fisher} or Vectors of Locally Aggregated Descriptors (VLAD)~\cite{jegou2010aggregating}. Learning place representations using deep networks has boosted the capabilities and performance of place recognition. The architectures consist of a convolutional backbone followed by a pooling operation, such as max-pooling~\cite{MAC2016Tolias} or average-pooling~\cite{Spoc2015Babenko}. Radenovid \etal~\cite{GeM2017Radenovic} proposed a Generalized Mean Layer (GeM) that learns the norm of the pooling-operator, and thus generalizes max- and average-pooling. Arandjelovic \etal~\cite{NetVlad2015Arandjelovic} proposed NetVLAD, a deep architecture that also learns the VLAD clusters. MultiViewNet~\cite{MultiviewNet2019} and Warburg \etal~\cite{warburg2020mapillary} incorporate multiple views to improve retrieval performance. The Bayesian triplet loss~\cite{BTL2020Warburg} mirrors the triplet loss, but allows a network to embed images into Gaussian distributions rather than points, and thus propagate uncertainties to image retrieval. More similar to our work, Sourav \etal~\cite{garg2019semantic} explored extreme viewpoint changes by having query and database images from opposite directions. 

\subsection{6-DoF Visual Localization} 
Methods for camera localization have traditionally been classified as either structure-based or regression-based~\cite{zhousurvey}. \textbf{Regression-based} methods train a deep network to directly regress the camera pose from an input image. Some notable approaches are PoseNet~\cite{PoseNet2015Kendall}, that estimates the absolute pose of a camera with respect to a scene, and the works by Laskar \etal~\cite{Laskar2017CVPRW} and Balntas \etal~\cite{Balntas_2018_ECCV}, that estimate the relative pose between two cameras. However, recent evaluations (Zhou \etal~\cite{zhou2020learn} among others) seems to show that direct pose regression is less accurate than the more traditional one based on feature extraction and matching.

\textbf{Structure-based} methods, on the other hand, predict the pose of the camera by matching features between a 3D model and 2D query images. 
Traditional handcrafted descriptors struggle to match images taken under strongly differing viewing conditions. Thus, modern localization methods rely on convolutional neural networks to extract features that are more robust to appearance and viewpoint changes. SuperPoint~\cite{Superpoint2017DeTone} consists of a convolutional encoder followed by two heads: one for classifying if a pixel is an interest point, and the other to encode a feature descriptor. D2Net~\cite{D2Net2019Dusmanu} has a single CNN that extracts dense features that serves both as descriptors and detectors. LOFTR~\cite{sun2021loftr}, on the other hand, takes a pair of images as input and via a ViT~\cite{VIT2020Dosovitskiy}-based transformer architecture estimates both keypoints and matches simultaneously. Another line of research has focused on learning local descriptors using image level supervision only~\cite{Delf2016iccv_noh, How2020eccv_tolias, NCN2018NeuripsRocco, superncn2020Kurzejamski}. 
DELF~\cite{Delf2016iccv_noh} learns a spatial attention that is used to pool the feature map and can thus be optimized similarly to retrieval networks, but via the attention mechanism yields local features. Combining networks that predict both a coarse place descriptor and local descriptors~\cite{sarlin2018cvpr, delg2020cao, patchnetvlad2021cvpr, yang2020ur2kid, lit2022icml} have shown to improve both efficiency and robustness. 

\subsection{Visual Localization Datasets} 
Many large localization datasets have been proposed in recent years. These datasets have particularly focused on viewpoint and appearance changes. Among the most relevant and used \textbf{place recognition datasets} are Nordland~\cite{nordland} with seasonal changes, Tokyo24/7~\cite{torii201524} with day-night changes, and MSLS~\cite{warburg2020mapillary}, which is currently the largest and most diverse place recognition dataset including viewpoint, structural, seasonal and day-night changes. 

\textbf{6-DoF datasets} have higher localization accuracy than place recognition datasets. The poses of these datasets are either obtained with SfM reconstructions or differential GPS that provides localization accuracies within $5$ cm. Oxford Robotcar~\cite{RobotCarDatasetIJRR} consists of a car-mounted camera that traverses the same loop $100$ times during a year in varying weather and day/night conditions, Extended CMU Seasons dataset~\cite{Sattler2018CVPR, Badino2011} is similarly recorded with a car-mounted camera. Aachen Day-Night~\cite{Sattler2018CVPR, Sattler2012ImageRF} consists of images from hand-held devices and focuses on day-night changes. 
ETH-Microsoft~\cite{eth_ms_visloc_2021} is a recent dataset for visual localization that, beside challenging day/night appearance changes, also covers indoor environments.
All these datasets only contain street-level images taken from a camera mounted on a vehicle or a handheld device. In contrast, our DAG dataset contains images taken by a ground vehicle and an airplane.

\begin{figure*}[hpt!]
    \centering
    \includegraphics[width=\linewidth]{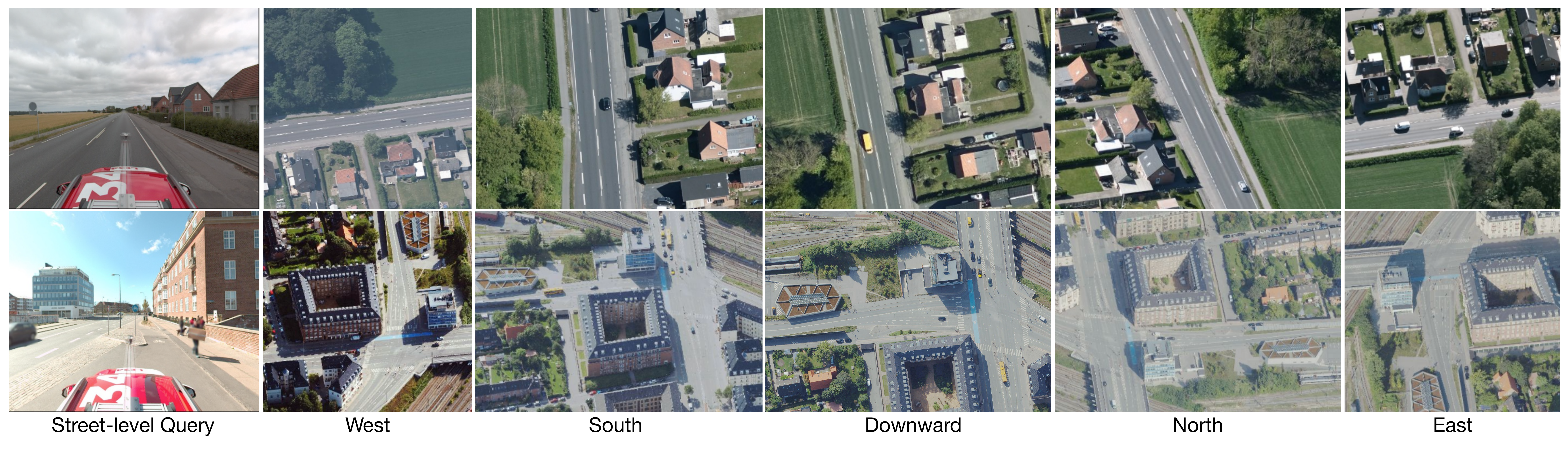}
    \vspace{-0.8cm}
    \caption{The airplane has five cameras mounted. One facing downwards and the rest facing each of the world corners (West, South, North, East). The figure shows images from the suburban environment in the \textit{Lolland} sequence and the harbor environment at the \textit{Nordhavn} sequences. 
    }
    \label{fig:dataset_examples}
\end{figure*}

\subsection{Aerial-to-Street-Level Retrieval and Localization} 

Aerial-to-Street-Level registration was addressed by Shan \etal~\cite{shan2014accurate}. Similar to us, they propose a view-dependent feature matching process. However, they assume to know the approximate position (within $20$ meter) of the street level image, while we first run a place recognition model to obtain this coarse localization. Another difference to their work is that their 3D reconstruction is created from the street-level image, which is only possible when multiple street-level images of the same area are available. We show that we can create an accurate 3D model from aerial images and render street-level images. This generalizes to scenes with only few street-level images of the scene.

Lin \etal~\cite{lin2015learning} propose to train a place recognition network for direct aerial-to-street-level retrieval. They construct a dataset that covers several large cities with both aerial and street level images. They train a place recognition network to be invariant to the extreme viewpoint change between aerial and street level images. In contrast to their work, we seek to find local correspondences to improve the coarse localization estimate of the place recognition model.

Most similar to our dataset is \cite{majdik2015air} that released a $2$ km sequence dataset captured by a drone in Zurich as well GoogleMaps images. Our dataset is much larger, covering more than $50$ km in more diverse urban and suburban environments. Their ground-truth car poses are based on GPS, and have as a consequence limited accuracy. In contrast, our street-level images have associated differential GPS and thus much higher pose accuracy.






\section{The Danish Airs and Grounds Dataset}

The DAG dataset contains aerial and street-level images from urban, suburban and rural regions in Denmark\footnote{The access to the data was possible thanks to the open access policy of the Danish Government, c.f.\@ \url{https://dataforsyningen.dk/}.}. The airplane photos are taken by five cameras; one facing vertically downward, and four oblique views facing each of the world corners (East, West, North, South). The images were recorded by The Danish Agency for Data Supply and Efficiency in 2017 and 2019. See Fig. \ref{fig:dataset_examples} for examples or visit their website for an interactive look at the images\footnote{\url{https://skraafoto.kortforsyningen.dk}}. The airplane photos poses are in principle within $5$ meter precision, which is further improved by visual alignment and reconstruction of multiple views. 

\begin{figure}[hpt!]
    \centering
    \includegraphics[width=0.9\linewidth]{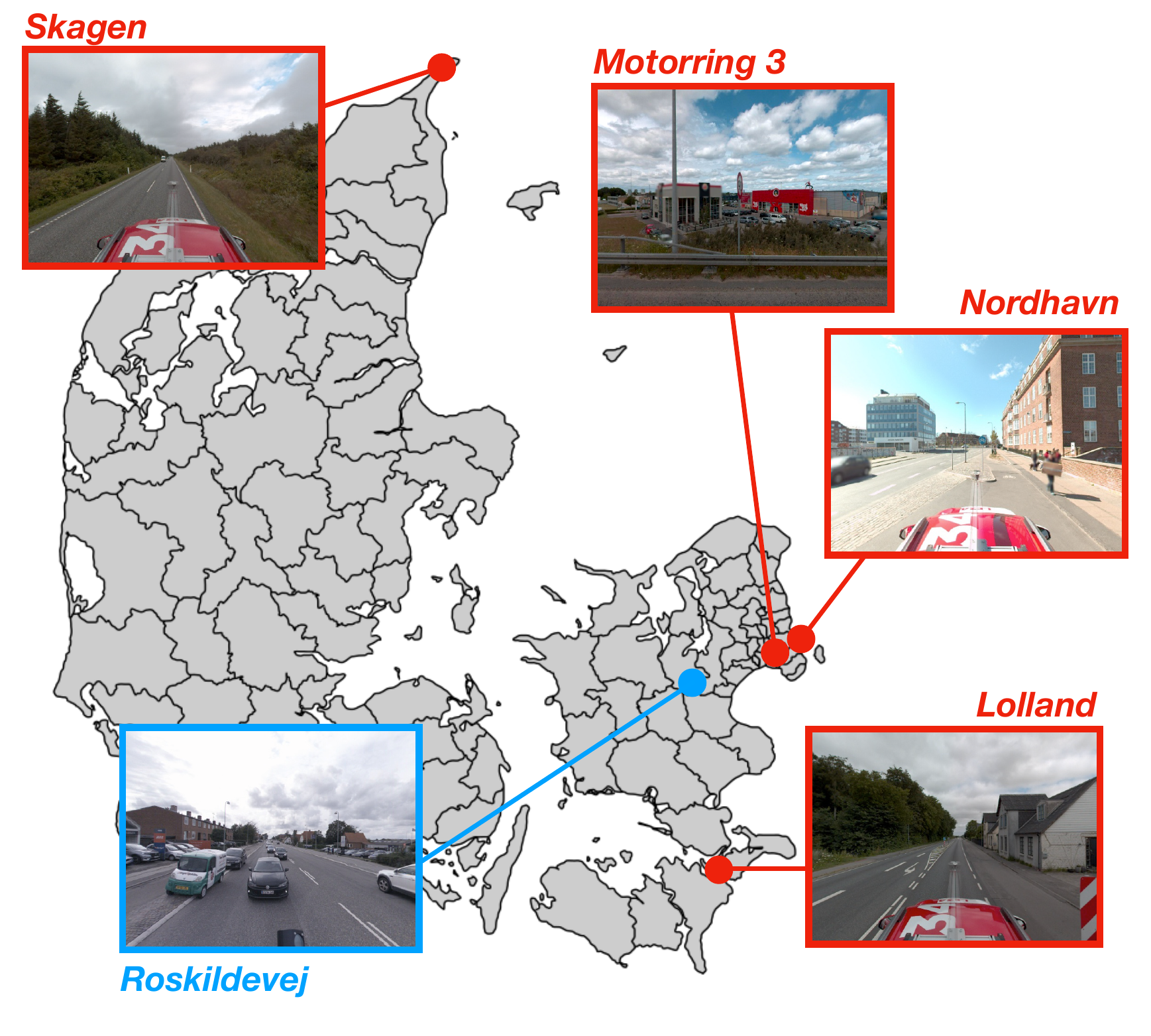}
    \caption{The five DAG sequences cover a large geographical area in Denmark. The sequences are captured at urban, suburban and rural regions, as highlighted with an example image from each sequence. The four red sequences are used for the training set and the blue sequence kept as the test set.}
    \label{fig:map_overview}
\end{figure}

\begin{figure*}[ht!]
    \centering
    \includegraphics[width=1\textwidth]{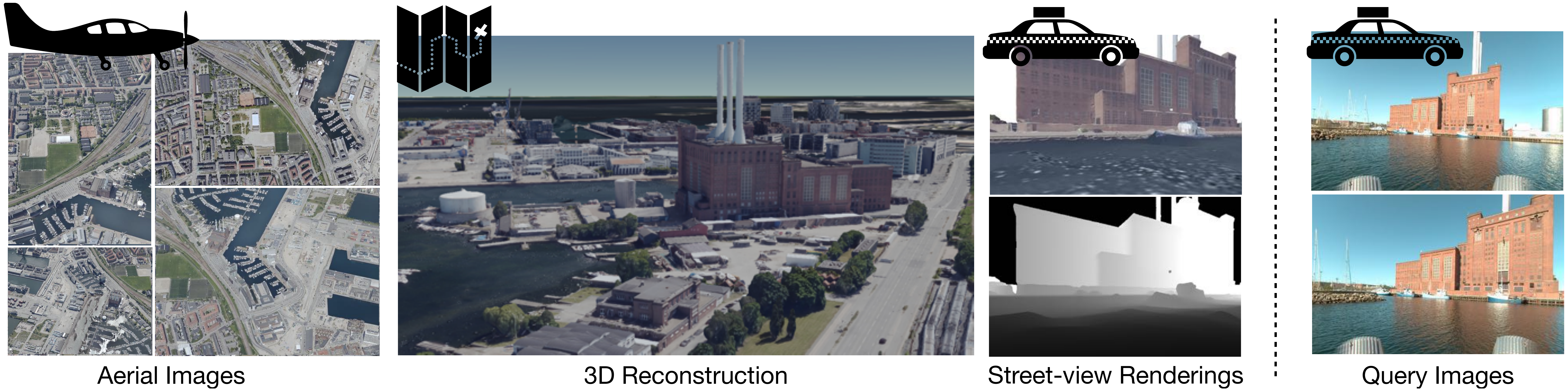}
    \caption{Street-level to aerial localization pipeline. We generate a dense and accurate multi-view 3D reconstruction from aerial images. Using this 3D model we can render a database of photo realistic images for real-image-to-render retrieval from actual street-view queries. Our experiments show that this is an effective pipeline for air-to-ground retrieval and localization.}
    \label{fig:overview_fig}
\end{figure*}%

The street-level images are recorded with a Ladybug5+ by the Danish Road Directory. Fig. \ref{fig:map_overview} shows the location of the five sequences; the \textit{Nordhavn} sequences consist of three sequences recorded in an urban harbor environment from a boat and a car. The \textit{Motorring 3} sequence is a suburban road around Copenhagen, the \textit{Roskildevej} sequence is in a urban environment, and both the \textit{Skagen} and \textit{Lolland} sequences are from rural areas in Denmark. The recorded sequences cover more than $50$ km and have more than $11,000$ panoramic images, which we project into four perspective cameras, totaling $44,000$ images. Fig. \ref{fig:map_overview} shows some examples of street-level images from different environments. The street-level images are labeled with global positioning metadata, which was obtained with a differential GPS unit with approximately $5$ cm accuracy.

\section{Street-level-to-Aerial Localization Pipeline}


We propose a localization pipeline that can be denoted as image-to-render, an intermediate category between image-to-image and image-to-map matching (following the terminology of \cite{williams2008image}). The extreme parallax angle and scale change between aerial and ground-level images renders image-to-image matching very challenging. Estimating intermediate 3D representations and rendering synthetic images at ground-level allows us to bridge viewpoint challenges and leverage all the recent deep models for image-to-image matching. Our experiments show that the appearance differences between real and rendered images are not an issue when matching deep features.

Our method consists of the following steps, which are also depicted in Fig. \ref{fig:overview_fig}: \textit{First}, we create a 3D model from the aerial images. \textit{Second}, we render street-level images from this 3D model in a regular grid. \textit{Third}, we use a place recognition method to retrieve street-level renderings from the same place as a given street-level query image. \textit{Fourth}, we use a structure-based localization method between the retrieved rendered image and the query image to obtain the $6$-DoF pose of the query image. 

\subsection{3D Reconstruction and Ground-Level View Synthesis}

We use the commercial software Agisoft Metashape\footnote{\url{https://www.agisoft.com/}} to create an accurate and dense 3D reconstruction from the aerial images. Due to the large computational and memory footprint of the 3D models, we partitioned each of the five sequences into sub-models of approximately $2$ kilometers. After that, we synthesize ground views of the 3D model in a regular grid with $5$-meter separation between synthetic cameras. We synthesize $8$ street-level renderings at each location at equally spaced directions ($45^{\circ}$ between each other). We set the intrinsics of the synthetic perspective cameras as the same as the camera used to record the query street-level images.




\begin{figure}
    \centering
    \begin{subfigure}{\linewidth}
        \centering
    \includegraphics[width=\linewidth]{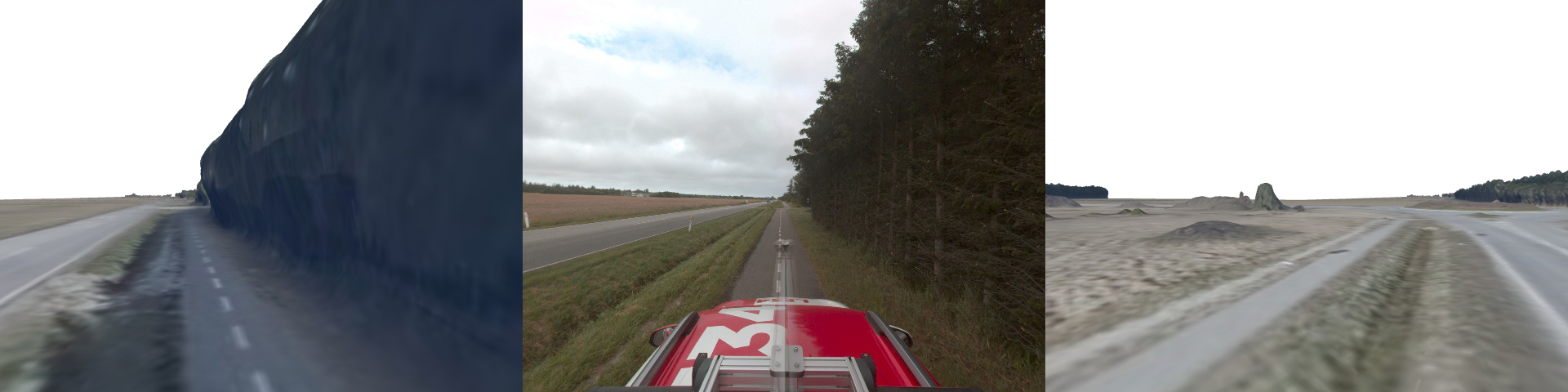}
    \end{subfigure}
    \begin{subfigure}{\linewidth}
        \centering
    \includegraphics[width=\linewidth]{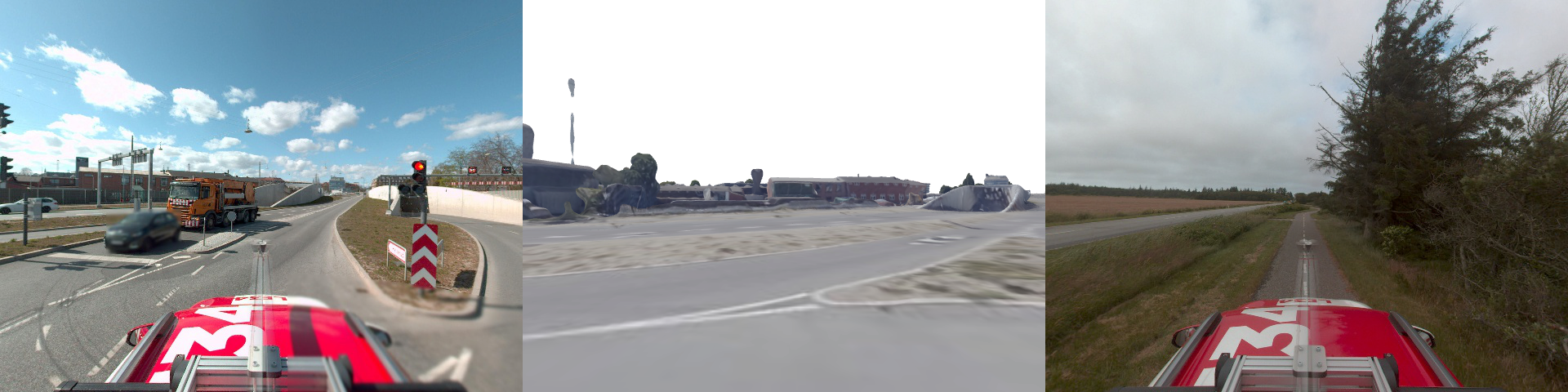}
    \end{subfigure}
    \begin{subfigure}{\linewidth}
        \centering
    \includegraphics[width=\linewidth]{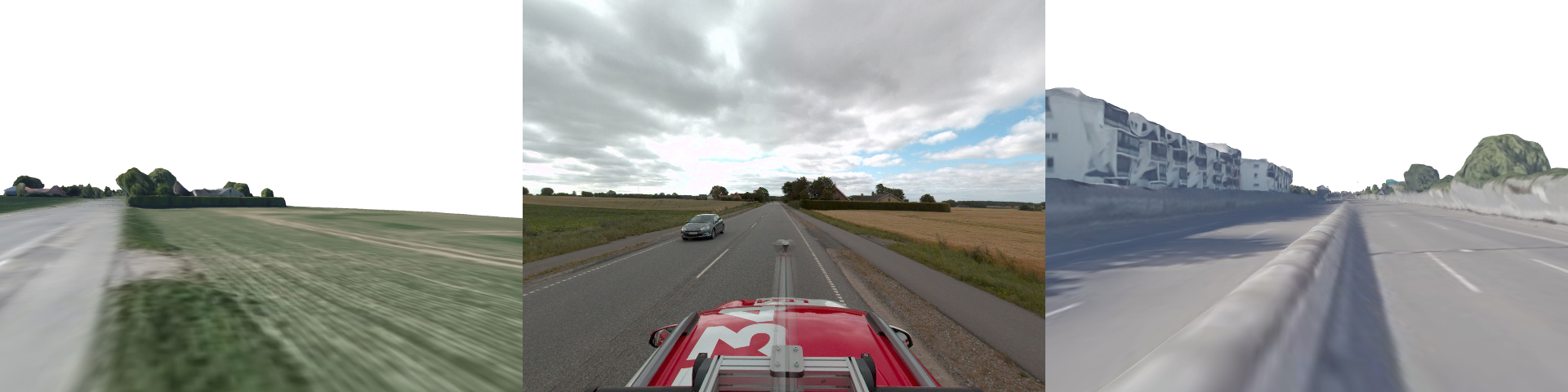}
    \end{subfigure}
    \begin{subfigure}{\linewidth}
        \centering
    \includegraphics[width=\linewidth]{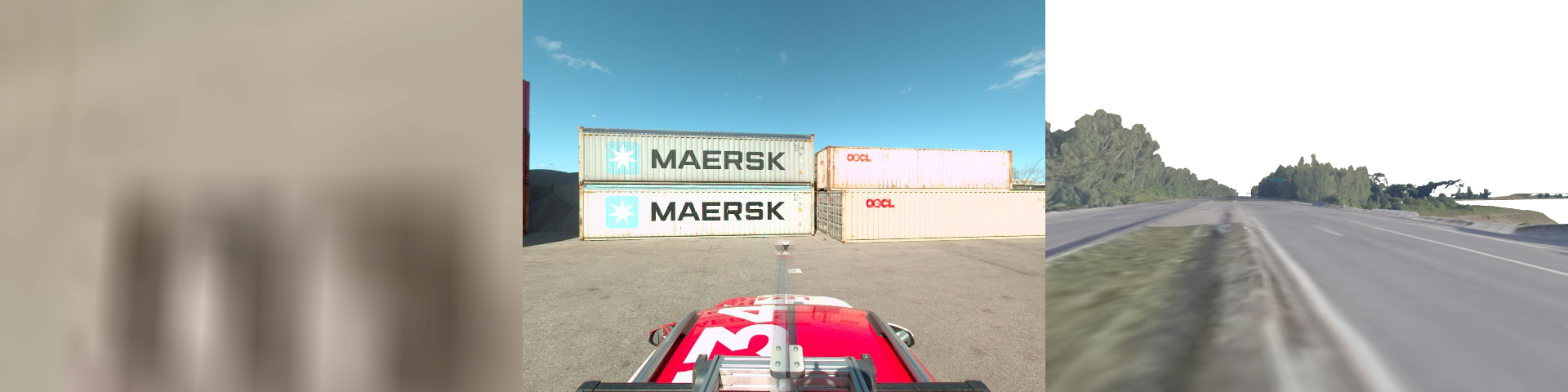}
    \end{subfigure}
    Anchor \quad \qquad \qquad Positive \qquad \qquad \quad Negative \\ \vspace{0.1cm}
    \caption{Examples of training triplets. The triplet loss will push the embedding of the positive closer to the anchor than the negative. Note that in the dynamic harbor environment, \textit{Nordhavn}, the containers are not in the same location when the car and the airplane visited the site. As seen in the last row, the rendering that is geographically close to the car, shows just the front of a container. These structural changes in the map makes visual localization very challenging.}
    \label{fig:triplets}
\end{figure}

\begin{table*}[!htp]
    \centering
    \begin{tabular}{l|rrrrrrr}
         &  R@1 & R@5 & R@10 & R@20 & M@5 & M@10 & M@20 \\ \hline \hline
        GeM         & 0.68 & 0.78 & 0.82 & 0.87 & 0.63 & 0.60 & 0.56  \\
        GeM (MSLS)  & \bf 0.80 & \bf 0.90 & \bf 0.92 & \bf 0.95 & \bf 0.77 & \bf 0.74 & \bf 0.71\\
    \end{tabular}
    \caption{Recall (R) and Mean Average Precision (M) at $\{1, 5, 10, 20\}$. Both methods consist of a ResNet50 followed by a Generalized Mean (GeM) layer. Pre-training on MSLS significantly improves the coarse localization performance.}
    \label{tab:retrieval}
\end{table*}

\subsection{Place recognition}
\label{sec:place_recognition_method}

We use a Resnet50 followed by the GeM aggregation layer~\cite{GeM2017Radenovic} as our place recognition network. We trained the network with the triplet loss and hard negative mining. We found that pre-training on the MSLS~\cite{warburg2020mapillary} significantly improves the retrieval performance. Fig. \ref{fig:triplets} shows examples of some of the triplets presented to the network during training. Note that the anchor and the positive are from the same place and the negative is from a different place. We found experimentally that it is important that the anchor and the positive image in each triplet are of the same type, either both synthetic or both real images.


\subsection{6-DoF Visual Localization}

Once the initial place is retrieved, our method proceeds with the actual 6-DoF localization, which is based on a Perspective-n-Point (PnP) solver~\cite{terzakis2020consistently}. The goal is to find the camera pose that, given a set of 3D points, minimizes the reprojection error of the 2D points in the camera plane. The peculiarity in our case is that the 3D points are calculated by back-projecting the pixels of the rendered camera with associated depth information, while the 2D projections are extracted from the original picture’s corresponding pixels.

We experimented with both SIFT~\cite{Lowe2004sift} and D2-Net~\cite{D2Net2019Dusmanu} as feature detectors and descriptors. We use the ratio test~\cite{Lowe2004sift} to filter matches for SIFT, but use a cross-matching check for D2-Net as suggested by the authors~\cite{D2Net2019Dusmanu}. We use only the $1000$ best matches to increase the chances for RANSAC convergence. Once the rendered picture matches were identified, each pixel in the rendered image was backprojected to obtain its 3D coordinates in the world using the depth of the 3D reconstruction. We then use a PnP solver to obtain the $6$-DoF pose between the 2D and 3D point correspondences. 

\section{Experimental Results}

In this section, we present the results of our proposed localization pipeline. We evaluate the place recognition method and $6$-DoF localization separately to establish baseline results for the two individual tasks.

\subsection{Place Recognition}

\begin{figure}[ht!]
    \centering
     \begin{subfigure}{0.49\linewidth}
         \centering
         \includegraphics[width=\linewidth]{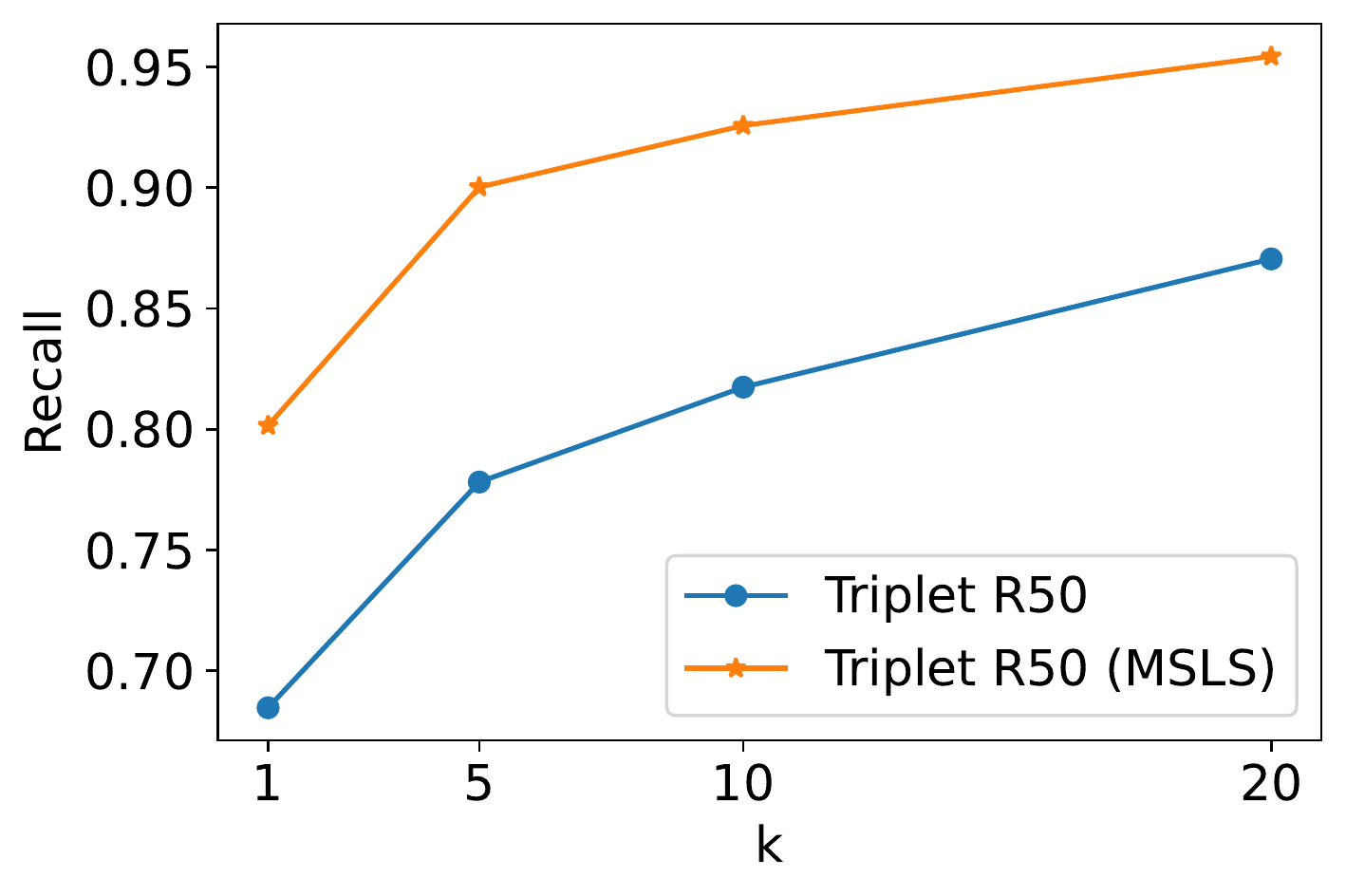}
     \end{subfigure}
     \begin{subfigure}{0.49\linewidth}
         \centering
         \includegraphics[width=\linewidth]{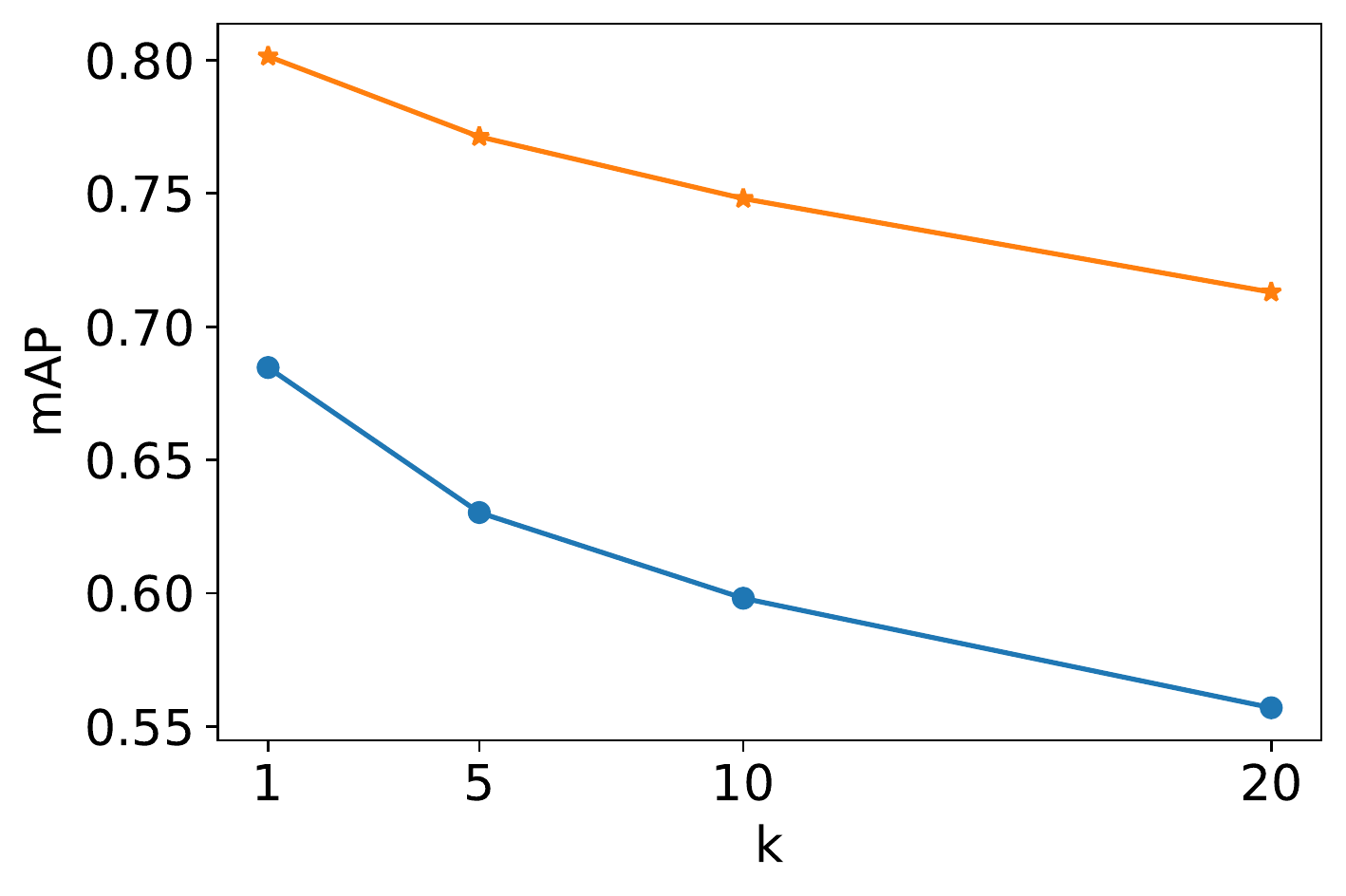}
     \end{subfigure}
    \caption{Recall and Mean Average Precision at $k$ for a Resnet50 with GeM pooling trained on DAG (Resnet50 triplet), and the same architecture first pre-trained on MSLS, and then fine-tuned on DAG (Resnet50 triplet (MSLS)).}
    \label{fig:recall}
\end{figure}

Fig. \ref{fig:recall} and Table \ref{tab:retrieval} show the mean average precision (mAP@k) and recall@k evaluated at $k$ number of nearest neighbors on the test sequence, \textit{Roskildevej}. A Resnet50 with a GeM-layer, trained with the triplet loss (Triplet R50 in the figure) correctly retrieves the same-place database image $69 \%$ of the times (Recall@1 is $0.69$) in the test set. The figure shows that the models can for most queries perform coarse localization by identifying the rendered images within a $25$ meter radius. Pre-training the network on the very large place recognition dataset MSLS~\cite{warburg2020mapillary}, and then fine-tuning on DAG, results in a significantly improved performance (Triplet R50 (MSLS) in the figure). With this setup, the Recall@1 increases to $0.80$.

\begin{figure}
    \centering
    \begin{subfigure}{\linewidth}
        \centering
    \includegraphics[width=\linewidth]{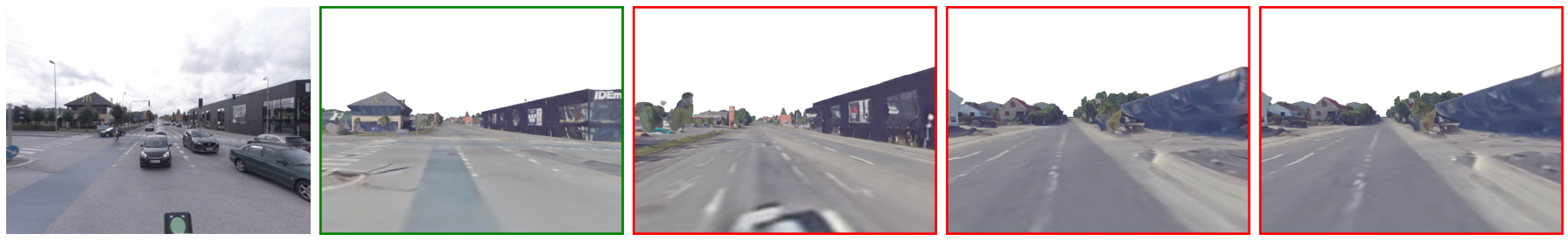}
    \end{subfigure}
    \begin{subfigure}{\linewidth}
        \centering
    \includegraphics[width=\linewidth]{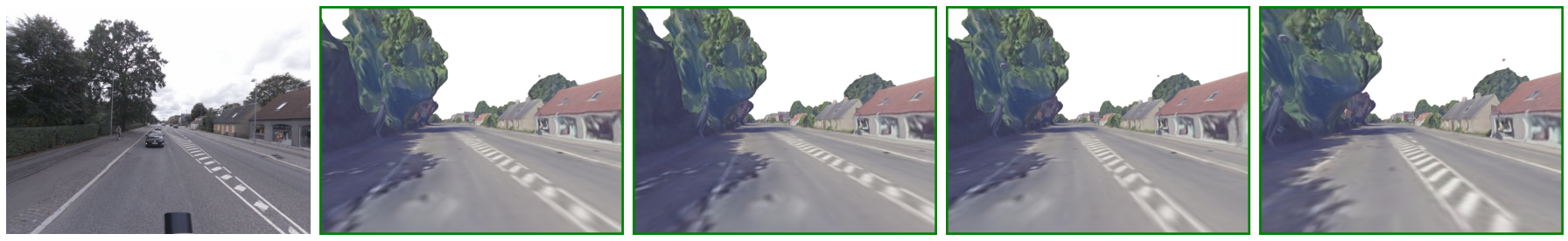}
    \end{subfigure}
    \begin{subfigure}{\linewidth}
        \centering
    \includegraphics[width=\linewidth]{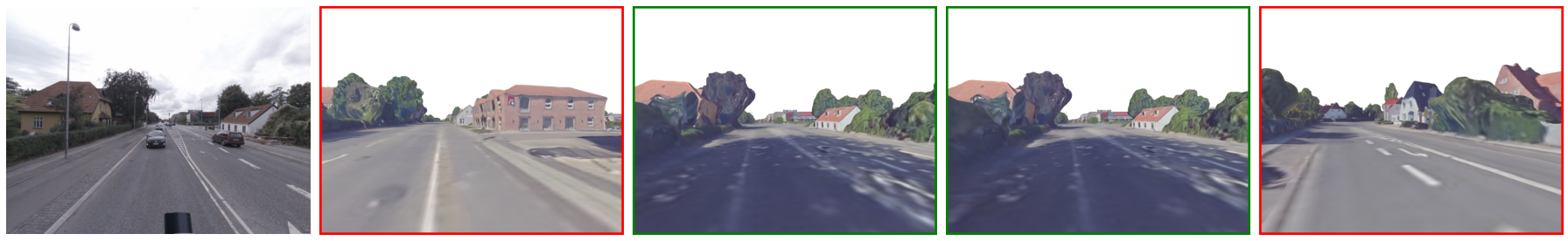}
    \end{subfigure}
    \begin{subfigure}{\linewidth}
        \centering
    \includegraphics[width=\linewidth]{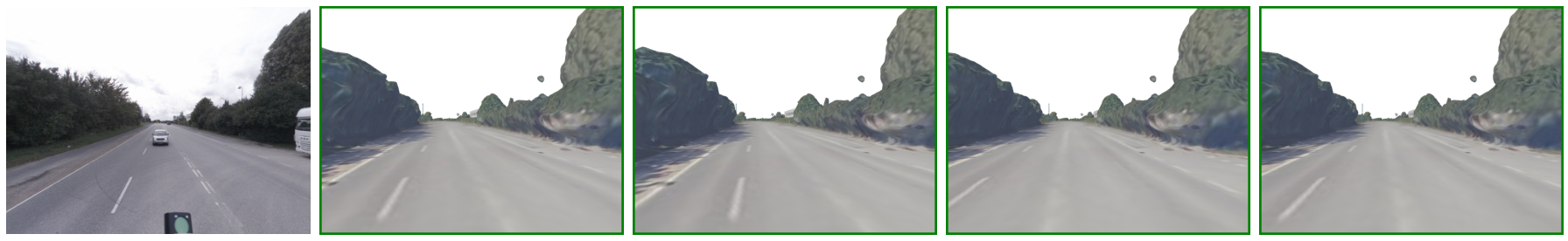}
    \end{subfigure}
    \begin{subfigure}{\linewidth}
        \centering
    \includegraphics[width=\linewidth]{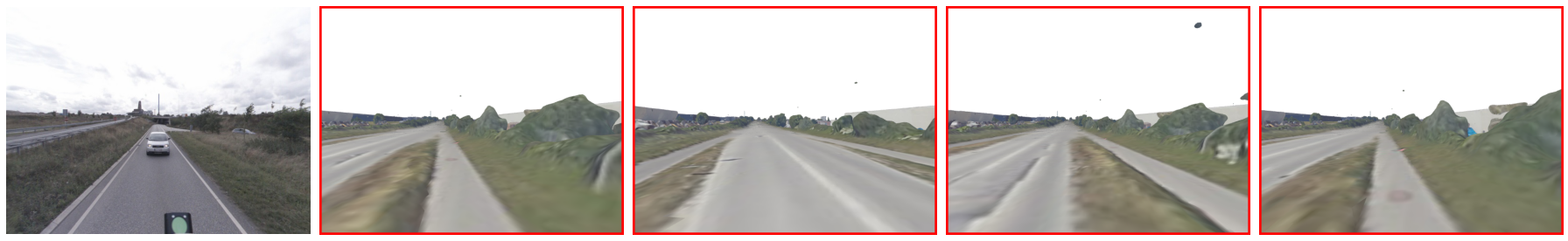}
    \end{subfigure}
    \caption{Qualitative retrieval results. The first column shows five query images, and the rest of columns show in order the four closest rendered images retrieved from the dataset. The green/red boundary indicates if the distance between the query and retrieved images is less (green) or greater (red) than $25$ meters. Observe that the network learned to be invariant to textural differences between rendered images and real queries. Our method is especially challenged in areas without buildings.}
    \label{fig:retrieval_qualitative}
\end{figure}

Fig. \ref{fig:retrieval_qualitative} shows some qualitative examples of the network retrievals. The network struggles in scenes with dynamic objects and vegetation. We believe that vegetation is a particularly challenging instance of this dataset. The aerial images and street-level images are not taken at the same time, thus trees and bushes change appearance (summer/winter). Furthermore, one of the limitations of our localization pipeline is that the 3D reconstruction of vegetation is very coarse. As the aerial images are not taken at the same time, changes in the vegetation (motion caused by the wind, vegetation growth or seasonal effects) result in a smoothing of the 3D reconstruction. Research into direct aerial to street-level localization (without 3D reconstruction) or learning methods that consider such changes are promising directions as they can circumvent this limitation. 

\subsection{Visual Localization}

In this section we evaluate the localization error of the relative pose between a query image and rendered image retrieved by the neural network as described in Section \ref{sec:place_recognition_method}. We consider two alternatives, using SIFT and D2Net features, and report translation and rotation errors. 

\begin{figure}[h!]
     \centering
     \begin{subfigure}{0.49\linewidth}
         \centering
         \includegraphics[width=\linewidth]{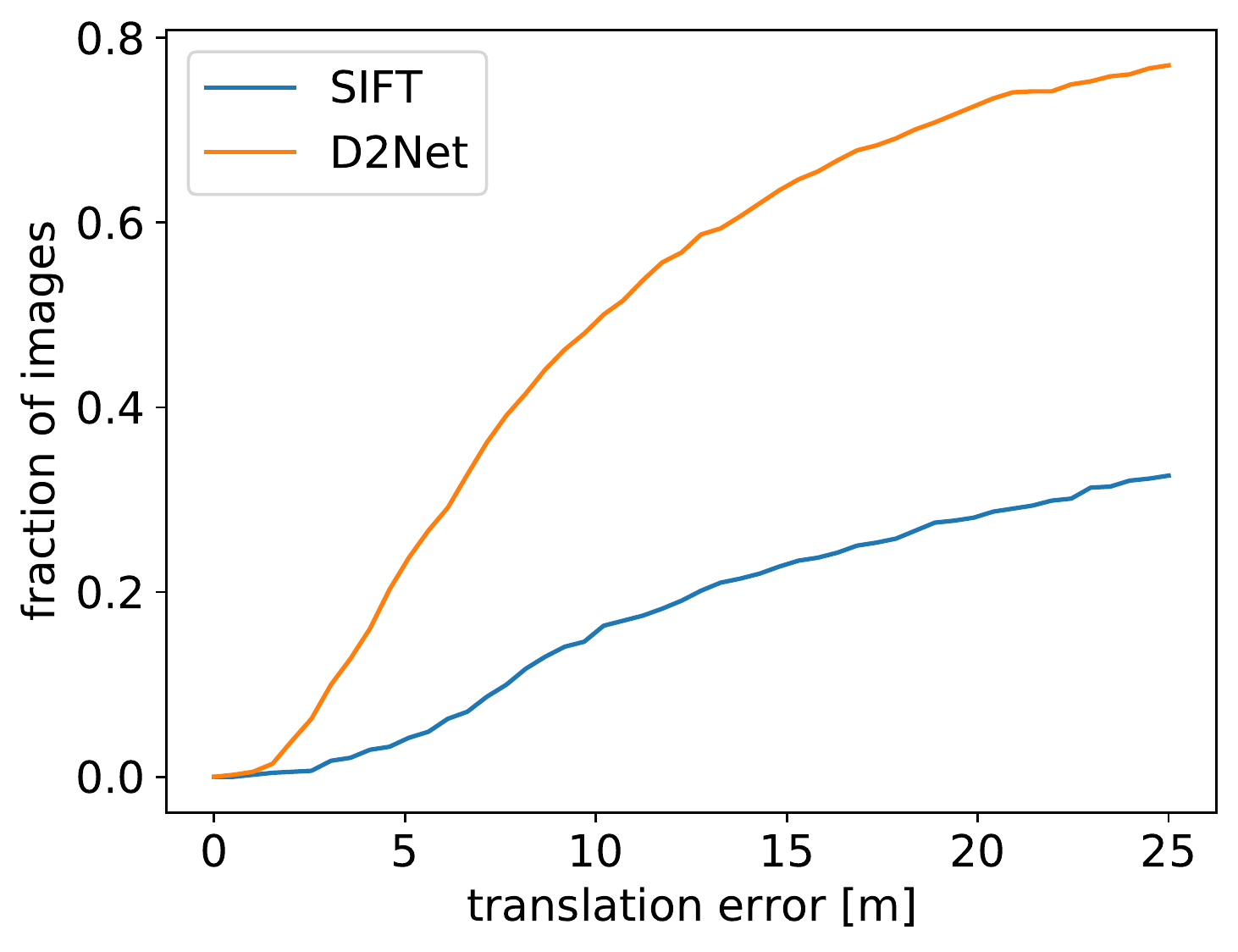}
     \end{subfigure}
     \begin{subfigure}{0.49\linewidth}
         \centering
         \includegraphics[width=\linewidth]{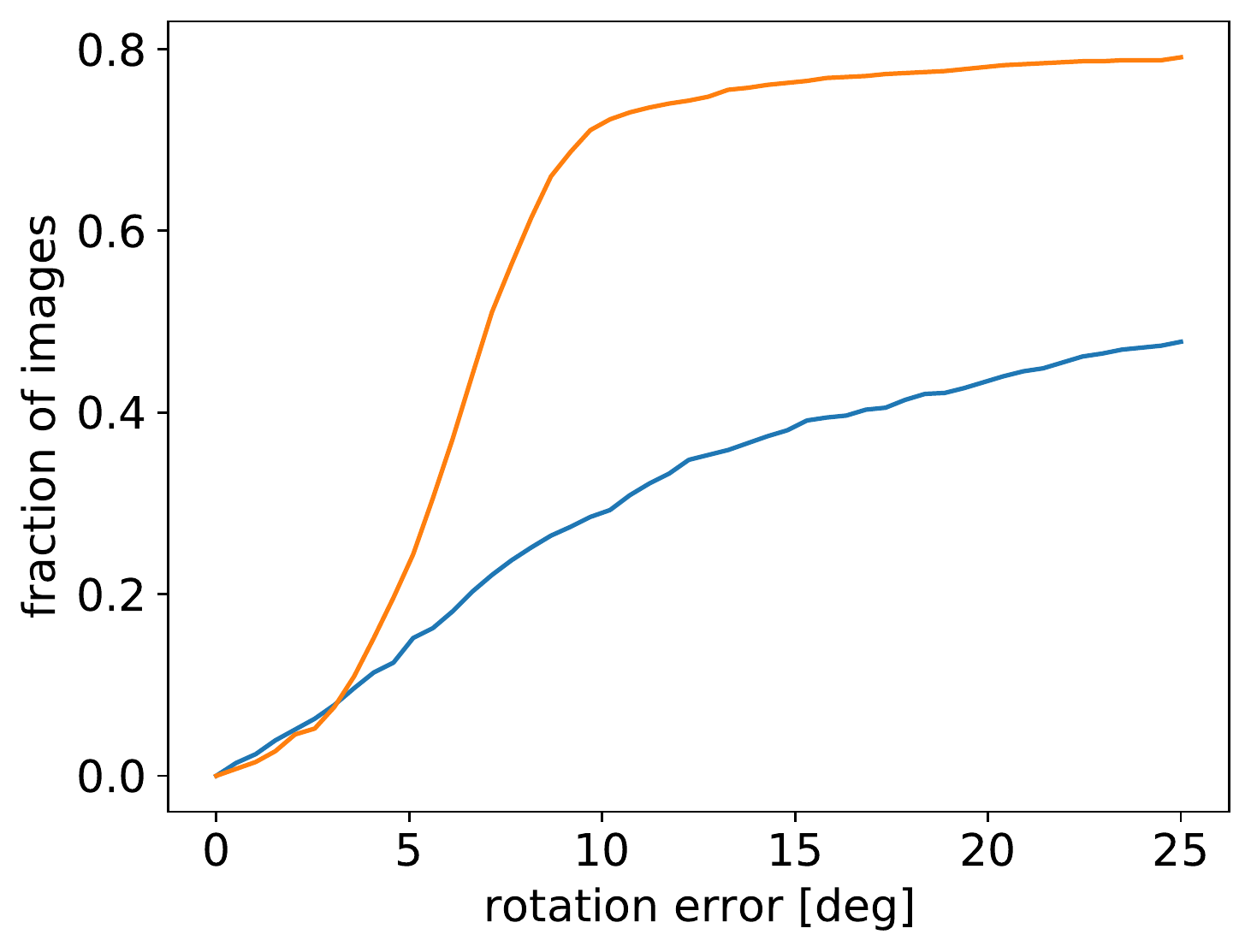}
     \end{subfigure} 
    \caption{Cumulative plots for translation and rotation absolute errors using SIFT and D2Net as feature detectors/descriptors. D2Net looks at higher level structures and thus generalizes between the different textures in the rendered and query images. It clearly outperforms the SIFT detector.}
    \label{fig:localization_accuracy}
\end{figure}

Fig. \ref{fig:localization_accuracy} shows cumulative error graphs (fraction of images with translational and rotational errors under different thresholds) for both cases. D2Net offer a significant improvement over SIFT. The reason is that SIFT describes the low-level textural appearance (which is very different between the query image and the rendered images), whereas D2Net has learned a higher level, more semantically meaningful description of the features, that is less dependent on specific low-level texture patterns.

\begin{figure}[!htp]
    \centering
    \begin{subfigure}{\linewidth}
        \centering
        \includegraphics[width=\linewidth]{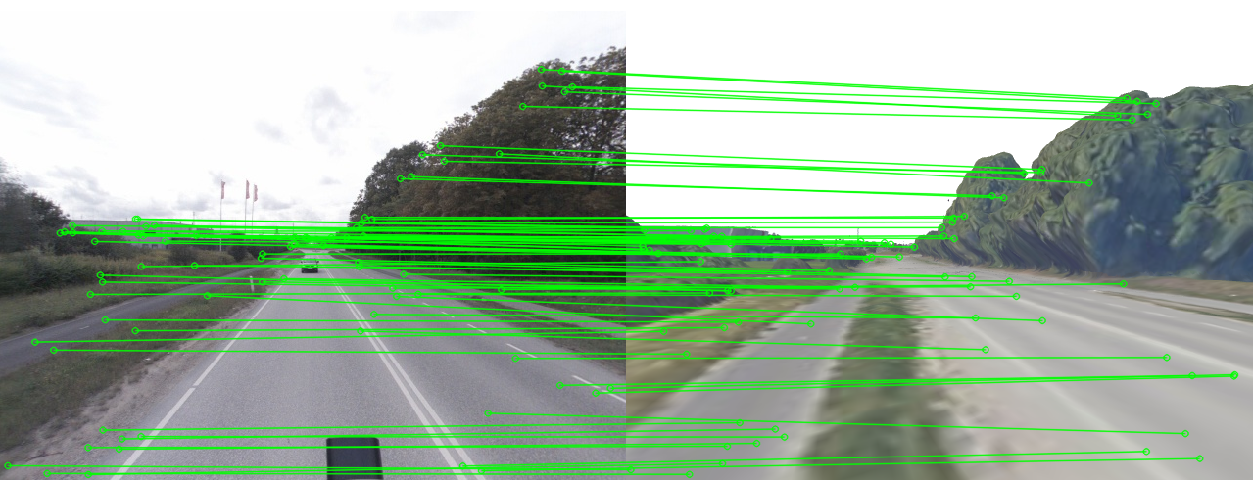}
    \end{subfigure} \\
    
    \begin{subfigure}{\linewidth}
        \centering
        \includegraphics[width=\linewidth]{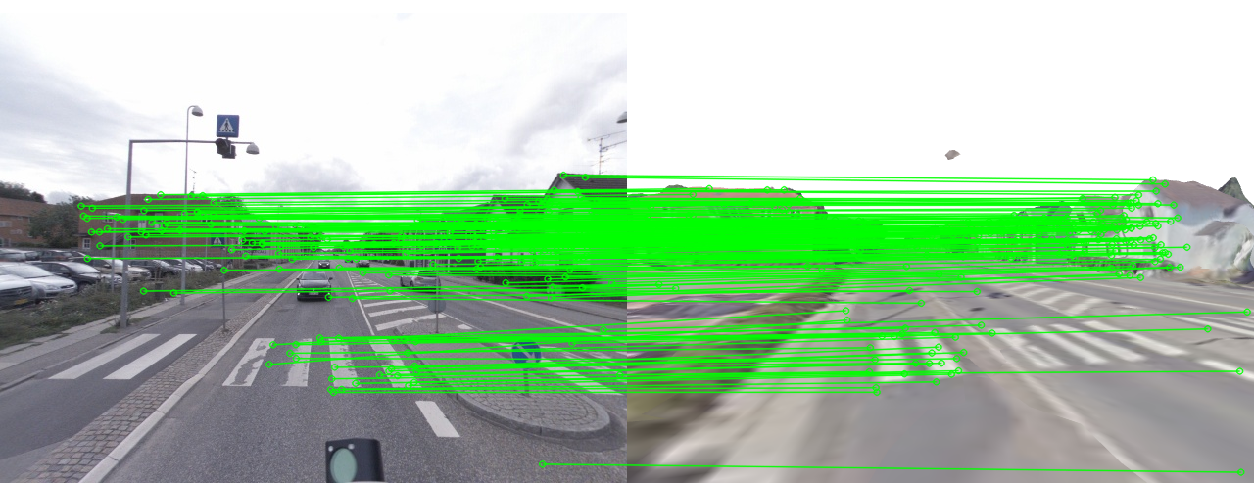}
    \end{subfigure}\\
    
    \begin{subfigure}{\linewidth}
        \includegraphics[width=\linewidth]{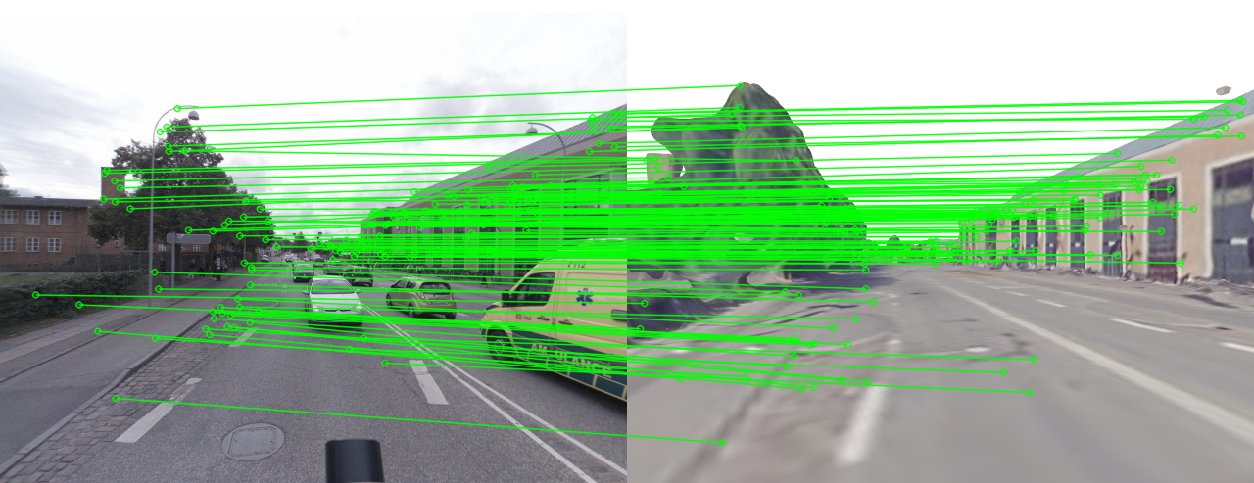}
    \end{subfigure}
    
    \caption{Inlier matches between query images and the street-level renderings. Since aerial images are taken at different times, moving objects such as cars and leaves are smoothed out or blurred. However, D2Net is still able to find reliable matches between the two views. The challenging aerial-to-street-level viewpoint changes are alleviated via 3D renderings and pre-trained feature detectors can be used.}
    \label{fig:qualitative_matches}
\end{figure}

Fig. \ref{fig:qualitative_matches} shows several examples of D2Net matches between the query and database images after cross matching. 
Observe how the features extracted on buildings have in general low image errors. Matches on the road and in vegetation, on the other hand, have a coarser localization in the image. 

\begin{table}[htp!]
    \centering
    \begin{tabular}{l|rrr}
         &  5m/5$^{\circ}$ & 10m/10$^{\circ}$ & 25m/25$^{\circ}$ \\ \hline \hline
        GeM (MSLS) + SIFT & 0.01 & 0.11 & 0.29\\
        GeM (MSLS) + D2Net & \bf0.04 & \bf0.42 & \bf0.66\\
    \end{tabular}
    \caption{Localization accuracy evaluated for all queries. Table shows the ratio of queries that satisfies both translation and rotation thresholds. Table shows that D2Net features outperforms SIFT features. }
    \label{tab:localization_accuracy}
\end{table}

Note in Fig. \ref{fig:localization_accuracy} that the median translation error is approximately $5$ meters, which roughly agrees with a quick geometric estimate. The aerial image resolution is $10$ centimeters per pixel. Assuming matching errors over $1$ pixel, parallax angles between $20^\circ$ and $40^\circ$ and small translation and rotation errors, they propagate to triangulation errors over $1$ meter. Such reconstruction errors may be bigger for textureless areas, vegetation and dynamic objects, and propagate to the localization via PnP. Our optimal RANSAC threshold is $40$ pixels, which indicates that there exist matches with high error that also add up to the localization error. We also observed unevenly distributed matches. Simulations of the geometry of the problem gave errors of the same level as those obtained with the real data.

Table \ref{tab:localization_accuracy} shows the ratio over \emph{all} retrievals with error under certain translation/rotation errors (e.g., 42\% of all queries have localization errors below 10m/10$^{\circ}$). Observe again the substantial difference between SIFT and D2Net.

\section{Additional Visualizations of the Data}

\begin{figure*}
    \centering
    \begin{subfigure}{0.3\linewidth}
        \centering
        \includegraphics[width=\linewidth]{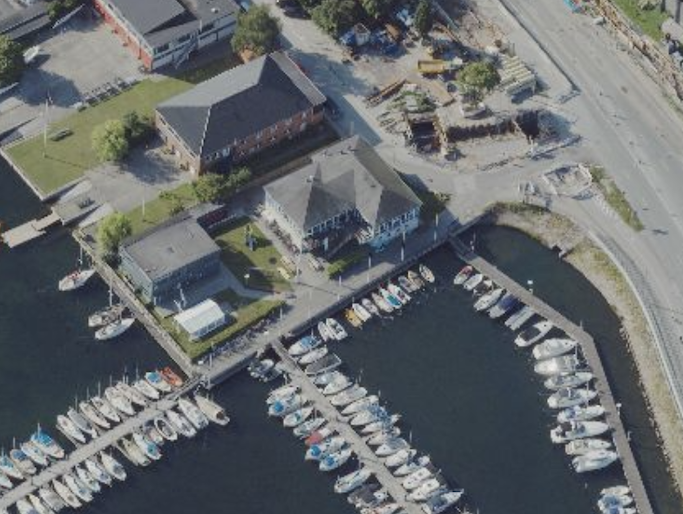}
    \end{subfigure}
    \begin{subfigure}{0.3\linewidth}
        \centering
        \includegraphics[width=\linewidth]{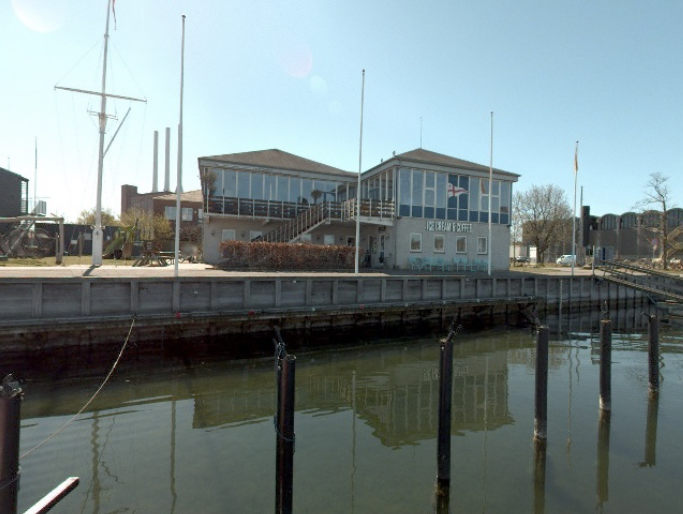}
    \end{subfigure}
    \begin{subfigure}{0.3\linewidth}
        \centering
        \includegraphics[width=\linewidth]{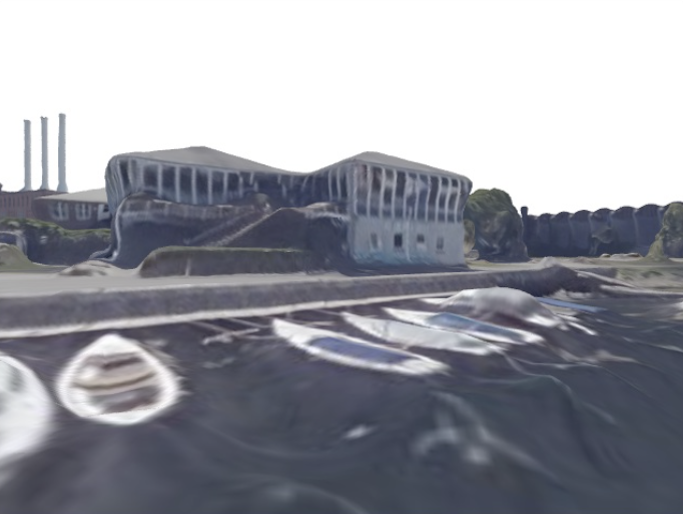}
    \end{subfigure} \\ \vspace{0.1cm}
    
    \begin{subfigure}{0.3\linewidth}
        \centering
        \includegraphics[width=\linewidth]{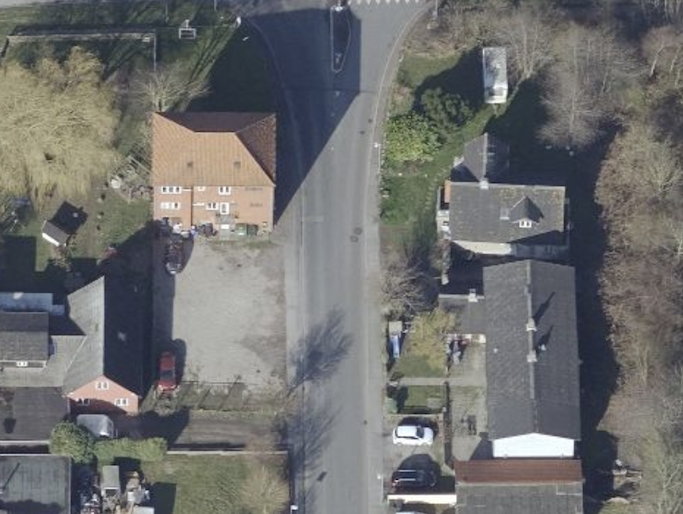}
    \end{subfigure}
    \begin{subfigure}{0.3\linewidth}
        \centering
        \includegraphics[width=\linewidth]{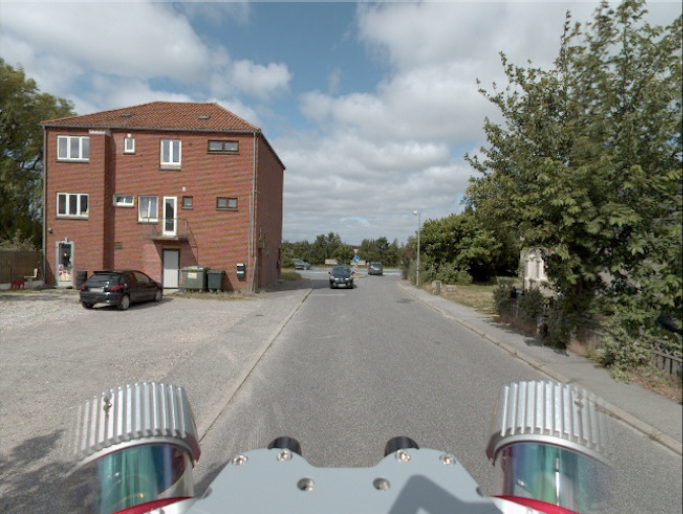}
    \end{subfigure}
    \begin{subfigure}{0.3\linewidth}
        \centering
        \includegraphics[width=\linewidth]{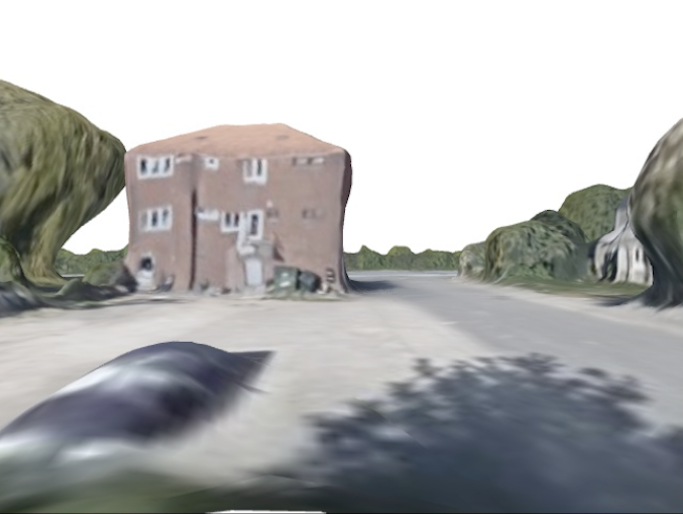}
    \end{subfigure} \\ \vspace{0.1cm}
    
    \begin{subfigure}{0.3\linewidth}
        \centering
        \includegraphics[width=\linewidth]{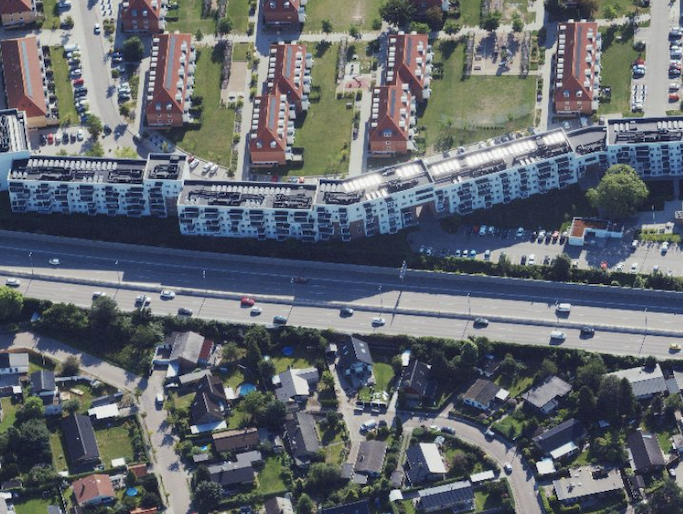}
    \end{subfigure}
    \begin{subfigure}{0.3\linewidth}
        \centering
        \includegraphics[width=\linewidth]{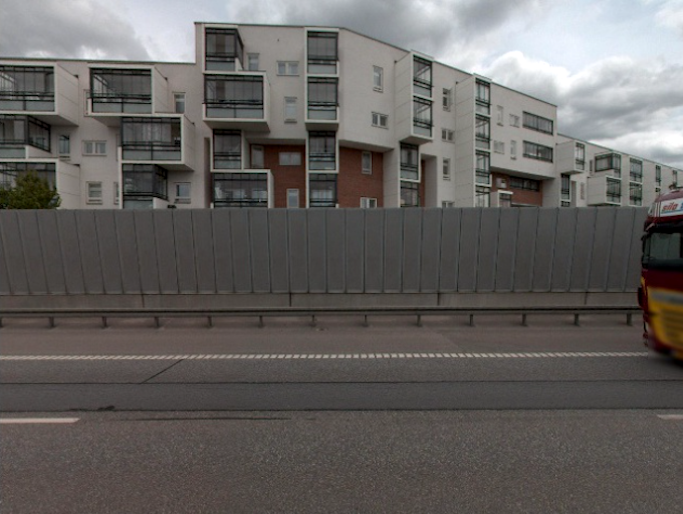}
    \end{subfigure}
    \begin{subfigure}{0.3\linewidth}
        \centering
        \includegraphics[width=\linewidth]{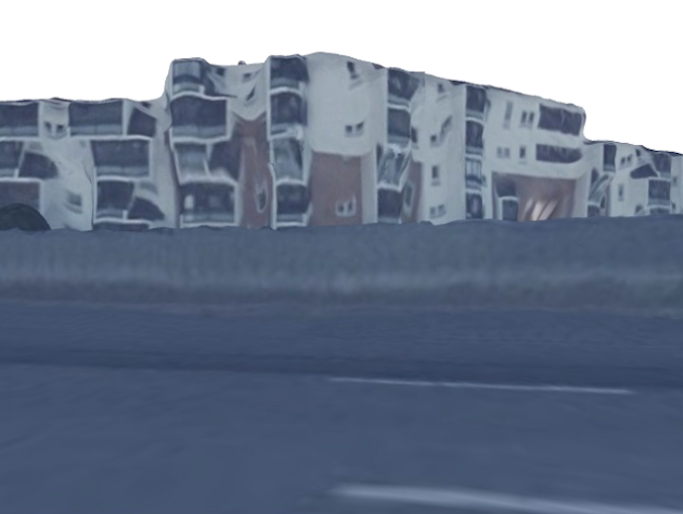}
    \end{subfigure} \\ \vspace{0.1cm}
    
    \begin{subfigure}{0.3\linewidth}
        \centering
        \includegraphics[width=\linewidth]{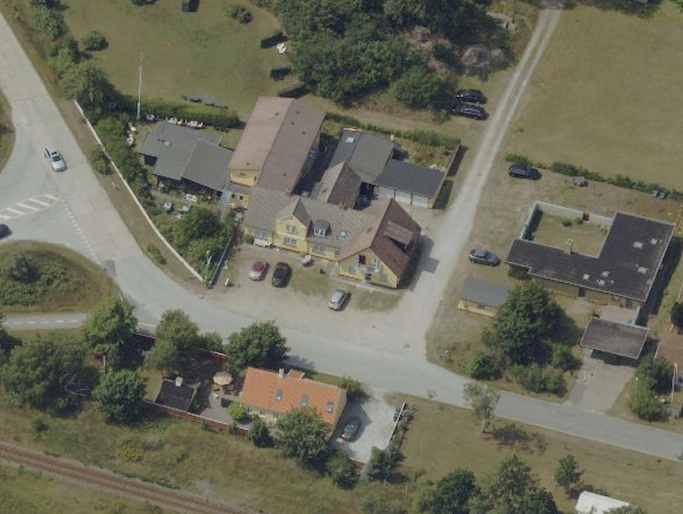}
    \end{subfigure}
    \begin{subfigure}{0.3\linewidth}
        \centering
        \includegraphics[width=\linewidth]{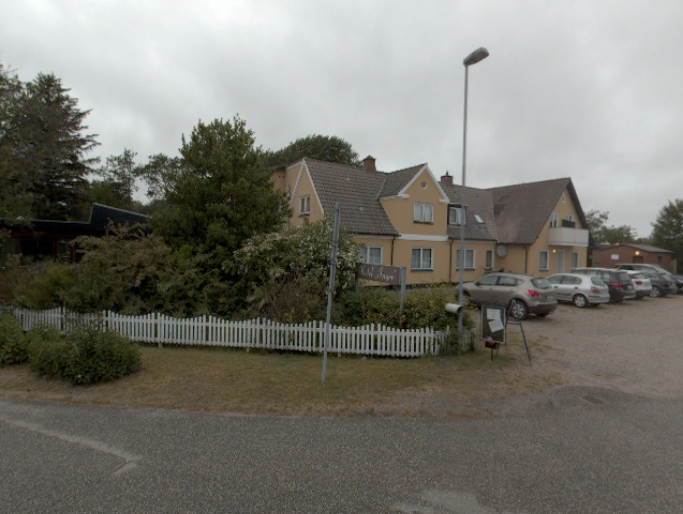}
    \end{subfigure}
    \begin{subfigure}{0.3\linewidth}
        \centering
        \includegraphics[width=\linewidth]{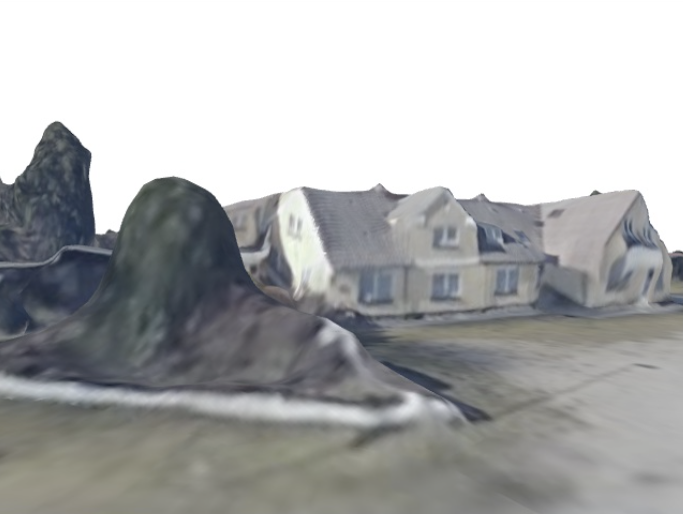}
    \end{subfigure} \\ \vspace{0.1cm}
    
    \begin{subfigure}{0.3\linewidth}
        \centering
        \includegraphics[width=\linewidth]{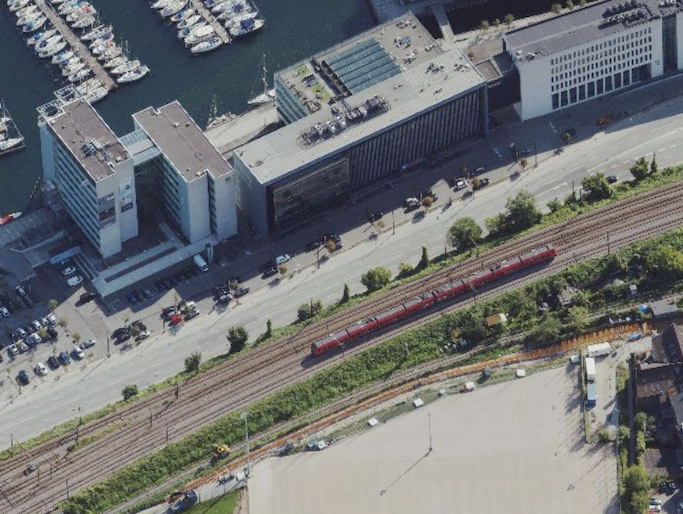}
        \caption{Aerial Images}
         \label{fig:air}
    \end{subfigure}
    \begin{subfigure}{0.3\linewidth}
        \centering
        \includegraphics[width=\linewidth]{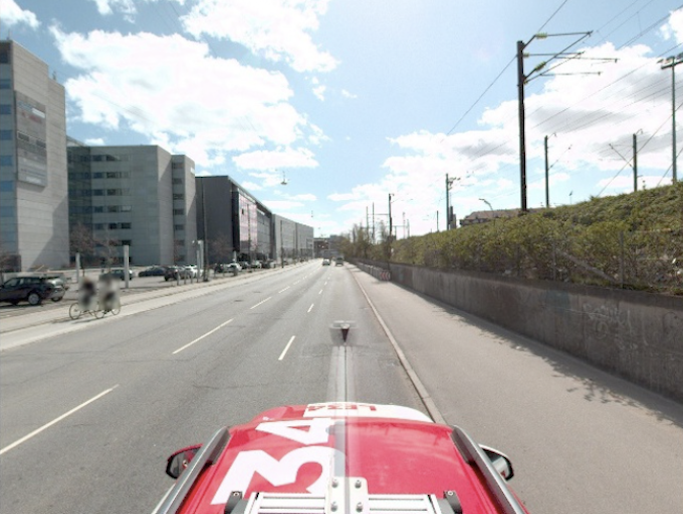}
        \caption{Street-View Images}
         \label{fig:query}
    \end{subfigure}
    \begin{subfigure}{0.3\linewidth}
        \centering
        \includegraphics[width=\linewidth]{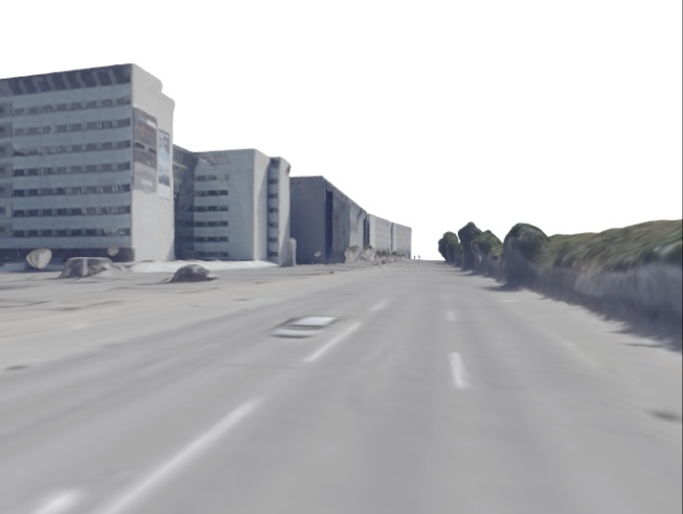}
        \caption{Street-View Renderings}
         \label{fig:databse}
    \end{subfigure}
    \caption{Sample visualizations of aerial, street-level images and street level renderings. The images illustrate the diversity of scenes in our DAG dataset.}
    \label{fig:more_viz}
\end{figure*}

In Figure \ref{fig:more_viz}, we present additional visualizations from the dataset. These images highlight again the difficulty of the problem and the diversity of the DAG dataset, covering urban, suburban and rural areas. As seen in the second row, the dataset also includes seasonal and dynamic changes between the aerial and street level image.

\section{Conclusions}

In this paper, we have presented \emph{Danish Airs and Grounds} (DAG), a dataset for aerial to street-level visual localization. Our data collection is the largest, up to date, that addresses such challenging setup. We believe there are two main aspects that make DAG relevant for the robotics and computer vision communities. Firstly, it addresses a particular case of wide baseline matching, which is one of the hardest cases for retrieval and localization. And secondly, from a more practical perspective, targets the relevant application case of street-level localization in aerial maps.

As a second contribution, we proposed a map-to-image re-localization pipeline for wide-baseline matching. In our experiments we analyze the performance of such an approach, serving as validation and initial baseline for our dataset.


{
\bibliographystyle{IEEEtran}
\bibliography{biblio}
}

\end{document}